# CF-JEPA: Mask-free forward prediction with asymmetric encoder utilization for time-series representation learning

**Jaehoon Lee[a], Sunghyun Sim [a b, *]**

[a]Graduate School of Artificial Intelligence Convergence Engineering,

Changwon National University, Changwon, South Korea

[b]Department of Artificial Intelligence Engineering,

Changwon National University, Changwon, South Korea

* Correspondence: ssh@changwon.ac.kr

**Abstract:** Self-supervised learning (SSL) for time-series representation learning is dominated by two paradigms: contrastive methods, which face challenges in constructing positive or negative pairs, and masking-based methods, which disrupt the temporal continuity of time-series signals. Joint-Embedding Predictive Architecture (JEPA) offers a promising alternative by predicting in representation space rather than reconstructing raw inputs. However, existing time-series JEPA variants still rely on masking and therefore inherit its continuity problem. Crop-based Forward JEPA (CF-JEPA) is proposed as an innovative mask-free framework that replaces masking with multi-horizon forward prediction: random crops serve as context views, and short-, mid-, and long-horizon future representations are predicted in the forward temporal direction, directly leveraging the inherent temporal ordering of time-series data as a learning signal. A strong asymmetry is also identified between the online encoder and the exponential moving average (EMA) target encoder, both produced from a single training run: the online encoder develops higher-rank discriminative features, while the EMA target encoder develops smoother, lower-rank temporal features. Exploiting this asymmetry, classification is routed to the online encoder and forecasting or anomaly detection to the EMA target encoder, achieving a 27% reduction in multivariate forecasting mean squared error (MSE) at no additional training cost. Across 126 University of California, Riverside (UCR) and 26 University of East Anglia (UEA) classification datasets, eight electricity transformer temperature forecasting benchmarks, and Key Performance Indicator /Yahoo anomaly detection, CF-JEPA achieves the highest average accuracy and rank on UCR and UEA among self-supervised baselines and ranks second on univariate forecasting and k-nearest neighbors-scored anomaly detection.

## 1. Introduction

Representation learning aims to train a model that automatically extracts latent representations useful for downstream tasks from raw data [1]. Among these, self-supervised learning (SSL) has emerged as a powerful paradigm for learning transferable representations from unlabeled data by exploiting inherent data structures. SSL has driven significant advances in natural language processing [2] and computer vision [3, 4, 5, 6], achieving performance competitive with supervised learning. Motivated by this success, SSL-based representation learning has also been actively investigated for time-series data as well [7, 8, 9].

Contrastive learning is one of the most widely adopted approaches for time-series representation learning [7, 8]. Contrastive methods learn representations by pulling positive pairs together and pushing negative pairs apart in the embedding space. Some approaches rely on data augmentation, such as scaling, warping, and permutation, to generate views [7, 10], whereas others use cropping to avoid augmentation [8, 9]. However, augmentation-based methods risk distorting temporal patterns. For example, scaling may alter significant amplitudes, warping may destroy frequency signatures, and permutations may break causal dependencies [11]. Furthermore, because contrastive methods define negatives based solely on the sample identity, different samples from the same class are pushed apart, potentially degrading the quality of the learned representations.

To avoid the difficulty of defining contrastive pairs, masking-based approaches [12, 13] have been actively studied for time-series SSL. Rather than relying on positive and negative comparisons, these methods hide portions of the input and train the model to reconstruct the masked content directly. Among these, Joint-Embedding Predictive Architecture (JEPA) [14] is a particularly promising framework because it predicts masked content in the representation space rather than reconstructing raw inputs. By operating in the representation space, JEPA can capture abstract high-level features without being constrained by low-level details of the raw signal [15, 16]. Existing JEPA implementations for both vision and time series [17, 18] generate a prediction target by masking parts of the input. Masking is well suited to images where spatial redundancy allows the model to infer hidden content from surrounding patches without disturbing the underlying visual structure. Time-series signals, however, lack this redundancy; masking individual time steps breaks the temporal continuity of the signal and destroys the patterns the model should learn to capture. The limitation, therefore, lies not in masking itself, but in masking that is structurally destructive when applied to temporally continuous signals.

To address this limitation, Crop-based Forward (CF)-JEPA is proposed, a mask-free framework that replaces masking with random forward cropping for view generation. Unlike masking, cropping preserves temporal continuity by extracting contiguous subsequences, keeping the underlying signal structure intact. CF-JEPA predicts future representations in the forward temporal direction for each crop, divided into short-, mid-, and long-horizon zones, maintaining JEPA's advantage of learning in the representation space without requiring masking, augmentation, or contrastive pairs. This design directly leverages the natural temporal ordering of time-series data as a learning signal, rather than artificially constructing prediction targets through input corruption.

Beyond the architectural proposal, the training dynamics of CF-JEPA were analyzed, and a strong asymmetry was identified between the online and exponential moving average (EMA) target encoders. Although both encoders are produced from a single training run, they developed markedly different representational properties: the online encoder retained higher-rank discriminative features suited to sample-level classification, whereas the EMA target encoder developed smoother, lower-rank temporal features suited to per-timestep regression. This asymmetry was exploited by routing classification to the online encoder and forecasting/anomaly detection to the EMA target encoder, achieving a 27% reduction in multivariate forecasting mean square error (MSE) at no additional training cost compared with single-encoder deployment. To our knowledge, this is the first time-series JEPA to explicitly identify and leverage dual-encoder asymmetry as a deployment time design choice. The main contributions of this study are as follows:

- **(C1)** CF-JEPA is introduced as a mask-free JEPA framework for time-series representation learning replacing masking with multi-horizon forward prediction, and eliminating the need for masking, augmentation, or contrastive objectives. Random forward crops serve as context views, and short-, mid-, and long-horizon future representations are predicted in the forward temporal direction by directly leveraging the inherent temporal ordering of time-series data.

- **(C2)** The asymmetric encoder utilization in CF-JEPA is identified and exploited, where the online and EMA target encoders, jointly produced from a single training run, develop distinct representational properties suited to different downstream tasks. Routing classification to the online encoder and forecasting or anomaly detection to the EMA target encoder achieves a 27% reduction in multivariate forecasting MSE compared to single-encoder deployment, with no additional training cost. To our knowledge, this is the first time series in which JEPA has identified and leveraged this asymmetry.
- **(C3)** Through comprehensive ablations of regularization, predictor design, predictor capacity, encoder backbone, and encoder routing, we show that (i) forward prediction is preferable to masking-based alternatives within the JEPA framework; (ii) lightweight linear predictors are sufficient for time-series JEPA; and (iii) backbone–objective compatibility, not capacity alone, governs downstream performance. Across 126 UCR and 26 UEA classifications, eight electricity transformer temperature (ETT) forecasting benchmarks, and Key Performance Indicator (KPI)/Yahoo anomaly detection, CF-JEPA achieved the highest average accuracy (Avg Acc) and average rank (Avg Rank) on University of California (UCR) and University of East Anglia (UEA) among self-supervised baselines and ranked second on univariate forecasting and k-nearest neighbors (k-NN)-scored anomaly detection.

The remainder of this paper is organized as follows: Section 2 reviews the related work. Section 3 describes the proposed method. Section 4 presents the experimental results and analysis. Section 5 discusses the implications of this study. Section 6 concludes the study.

## 2. Related work

### *2.1 Contrastive methods*

Contrastive methods for time-series SSL fall into two groups based on the view-generation strategy: augmentation-based and cropping-based. Augmentation-based methods apply handcrafted transformations to generate views. TS-TCC [7] uses temporal and contextual contrasts with two augmentation views generated by jittering, scaling, and permutations. InfoTS [10] introduces an information-aware framework that adaptively selects augmentations based on mutual information. AutoTCL [9] learns augmentation transformations automatically using a parameterized augmentation network. TimeCLR [23] adapts the SimCLR framework to univariate time series by introducing dynamic time warping (DTW)-based augmentation, which preserves temporal structure while generating phase-shift and amplitude-change variants paired with InceptionTime as the feature extractor. However, augmentation-based methods face the fundamental risk of distorting the temporal patterns that define class identity [11], and the optimal augmentation strategy remains highly domain-dependent.

Cropping-based methods avoid augmentation by extracting contiguous sub-sequences as views. TS2Vec [8] learns hierarchical timestamp-level representations through cropping and timestamp masking, achieving strong performance across classification, forecasting, and anomaly detection. SoftCLT [19] extends this framework by introducing soft contrastive assignments at both the instance and temporal levels. T-Rep [20] learns timestep-level representations by jointly training time embeddings with a feature extractor, capturing fine-grained temporal dependencies, and improving robustness to missing data. CoST [21] learns disentangled seasonal trend representations via contrastive learning in the frequency domain, using only cropped views. TimesURL [22] introduces frequency-temporal augmentation and double Universums as hard negatives, jointly optimizing contrastive learning with time reconstruction to learn universal representations applicable to multiple downstream tasks.

Although these crop-based methods avoid the distortion problem, they still rely on contrastive objectives that require defining positive and negative pairs, which can lead to false-negative assignments and sensitivity to pair-construction strategies. CF-JEPA shares the cropping strategy with this line but replaces the contrastive objective with forward prediction in the representation space, eliminating the pair-construction problem entirely.

### *2.2 Masking-based methods*

Masking-based methods learn representations by hiding portions of the input and training the model to recover masked content. This approach was popularized in NLP by BERT [2] and adapted for vision by MAE [12]. In the time series domain, SimMTM [13] reconstructs a masked time series by aggregating information from multiple masked versions using series-wise similarities. TimeMAE [25] adapts an MAE-style masked reconstruction to time series with a Transformer encoder. PatchTST [24] applies patch-level masking with a Transformer backbone as a pretraining objective for time-series forecasting; it is included here for its masking-based pretraining strategy, while noting that it is primarily evaluated as an end-to-end supervised forecasting model rather than as a general-purpose representation learner. TempSSL [26] integrates Temporal Masked Modeling, which reconstructs future segments from historical context, with Temporal Contrastive Learning, which treats the context and targets as positive pairs to mitigate distribution shifts in time-series forecasting.

These methods do not require augmentation or contrastive pairs, which is an attractive advantage. However, they reconstruct the original input space, forcing the model to recover low-level details that may not be relevant to downstream tasks. Moreover, randomly masking individual time steps in a time series disrupts the temporal continuity, potentially destroying the patterns that the model should learn to capture.

### *2.3 Joint-embedding predictive architecture*

JEPA [14] addresses the limitations of raw space reconstruction by predicting masked content in the representation space. I-JEPA [15] applies this concept to images by encoding visible patches, predicting the representations of masked patches, and learning semantic features without handcrafted augmentation. V-JEPA [16] extends this to videos by predicting masked spatiotemporal regions. Several recent studies have adapted JEPA to time series. TS-JEPA [17] closely follows the I-JEPA paradigm by applying patch-based masking with a Transformer encoder. MTS-JEPA [18] targets multivariate time-series anomaly detection using a soft codebook for regularization. TF-JEPA [27] addresses biosignal analysis with a dual-encoder design operating across the time and frequency domains; note that TF-JEPA's "dual encoder" refers to two encoders processing two input modalities (time and frequency), which is distinct from CF-JEPA's dual-encoder asymmetry between an online and an EMA target encoder over a single input view. FEI [28] uses the JEPA paradigm with frequency-domain masking, constructing two inference branches that predict frequency-masked embeddings from the input series and vice versa, enabling noncontrastive representation learning without explicit positive/negative pairs.

However, all these approaches rely on masking, whether spatial, patch-based, or frequency-domain, without leveraging the natural temporal ordering of time-series data that can serve as an alternative learning signal. Furthermore, none of them analyze or exploit the representational asymmetry between the online encoder and the EMA target encoder that naturally arises during JEPA pretraining. Our work departs from this line of thought by replacing masking with forward prediction and identifying online/target asymmetry as a deployment-time design choice, directly leveraging the temporal structure of time-series data within the JEPA framework.

### *2.4 Non-contrastive regularization*

Non-contrastive methods learn representations without negative pairs by using regularization strategies that prevent representational collapse. BYOL [5] and SimSiam [29] use asymmetric architectures with stop-gradient operations and EMA target networks. Variance-Invariance-Covariance Regularization (VICReg) [6] takes a different approach by independently regularizing the variance, invariance, and covariance of the representations without requiring asymmetric designs or large batch sizes. Barlow Twins [30] drives the cross-correlation matrix of representations toward identity to achieve decorrelation. These regularization techniques were primarily developed and validated using computer vision.

Their application in time-series representation learning, particularly in combination with predictive architectures, remains under investigated. CF-JEPA uses VICReg as its regularization strategy within a forward-predictive JEPA framework, combining non-contrastive regularization with predictive learning for time-series data.

### *2.5 Comparison summary*

Table 1 summarizes how CF-JEPA differs from prior time-series SSL methods along four axes: view generation, learning objective, prediction space, and whether the temporal ordering of the signal is used directly as a learning signal. Among the listed approaches, CF-JEPA is the only one that combines crop-based view generation with JEPA-style prediction in the representation space while explicitly leveraging temporal ordering as the learning signal.

**Table 1.** Positioning of Crop-based Forward Joint-Embedding Predictive Architecture (CF-JEPA) against four families of time-series self-supervised learning (SSL) methods.

| Method | Family | View generation | Learning objective | Temporal ordering as signal |
|---|---|---|---|---|
| TS-TCC | Aug.-based contrastive | Augmentation | Contrastive | No |
| InfoTS | Aug.-based contrastive | Augmentation | Contrastive | No |
| AutoTCL | Aug.-based contrastive | Augmentation | Contrastive | No |
| TimeCLR | Aug.-based contrastive | Augmentation | Contrastive | No |
| TS2Vec | Crop-based contrastive | Cropping | Contrastive | No |
| CoST | Crop-based contrastive | Cropping | Contrastive | No |
| T-Rep | Crop-based contrastive | Cropping | Contrastive | No |
| SoftCLT | Crop-based contrastive | Cropping | Contrastive | No |
| TimesURL | Crop-based contrastive | Cropping | Contrastive | No |
| SimMTM | Masking-based reconstruction | Masking | Reconstruction | No |
| TimeMAE | Masking-based reconstruction | Masking | Reconstruction | No |
| PatchTST | Masking-based reconstruction | Patch masking | Reconstruction | No |
| TempSSL | Masking-based reconstruction | Temporal masking | Reconstruction | No |
| TS-JEPA | Time-series JEPA (masking) | Patch masking | JEPA prediction | No |
| MTS-JEPA | Time-series JEPA (masking) | Masking + codebook | JEPA prediction | No |
| TF-JEPA | Time-series JEPA (masking) | Time/freq. masking | JEPA prediction | No |
| FEI | Time-series JEPA (masking) | Frequency masking | JEPA prediction | No |
| CF-JEPA (ours) | JEPA (mask-free, forward) | Forward cropping | JEPA prediction | Yes |

## 3. Methodology

CF-JEPA consists of four components: (1) a view-generation module that produces random forward crops from the input time series, (2) an online encoder that maps crops to representations, (3) a set of horizon predictors that predict future representations in the forward direction, and (4) an EMA target encoder that provides prediction targets.

### *3.1 Problem Setup and Notation*

The problem of self-supervised representation learning for time-series data is considered. The goal is to learn an encoder that maps raw sequences to representations useful for diverse downstream tasks without access to labels during pretraining. Fig. 1. Overall framework of CF-JEPA. Stage 1 (Random Cropping): random crops are sampled as views from each series. Stage 2 (Forward Prediction Training): the online encoder maps each crop to a context representation, and a predictor predicts future zones in representation space. The EMA target encoder (stop-gradient) encodes the full series to provide targets. The loss function combines L1 prediction with regularization terms (variance, covariance, and multi-scale invariance). Stage 3 (Downstream Task): the online encoder is routed for classification, whereas the target encoder is routed for forecasting and anomaly detection (asymmetric utilization).

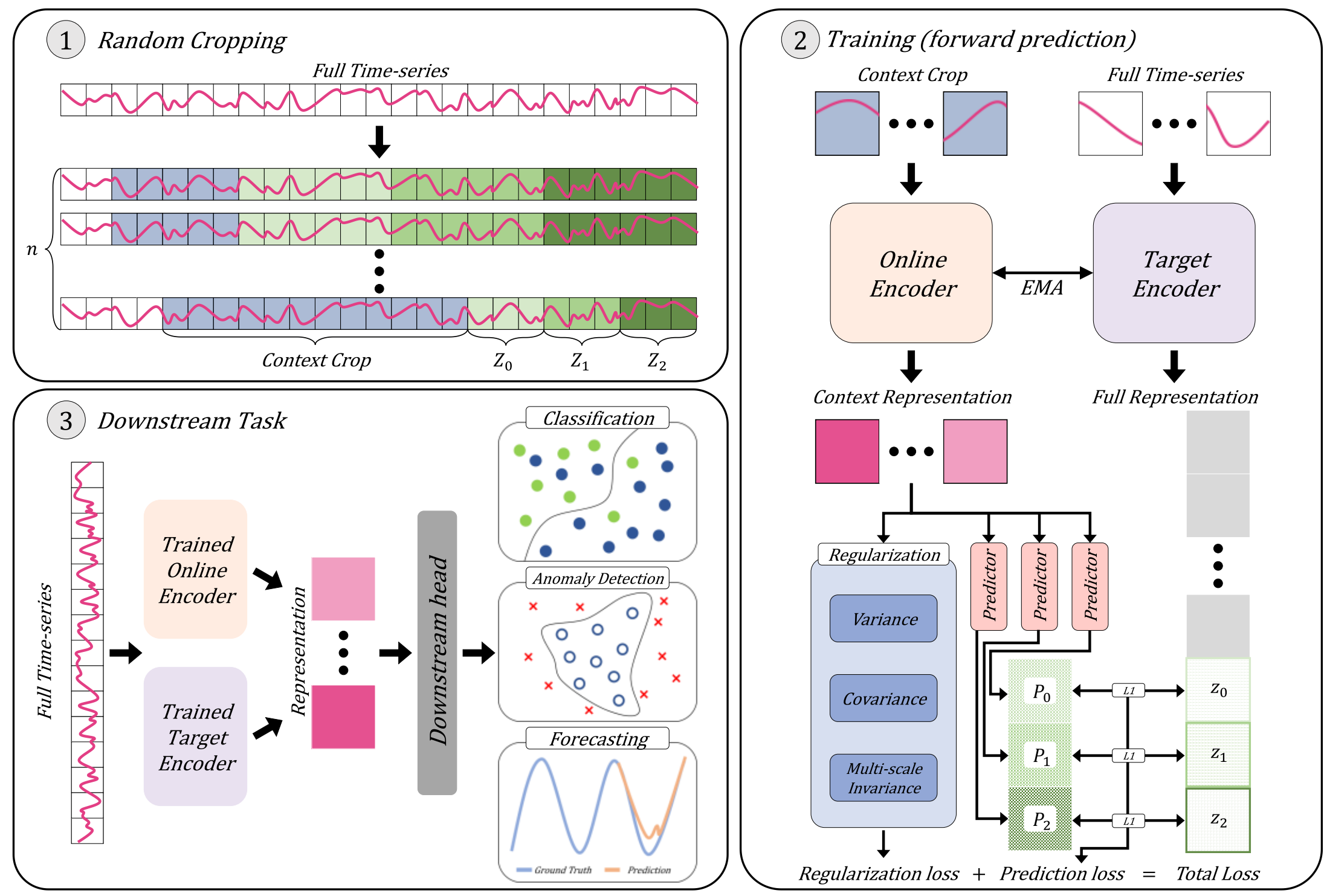

**Fig 1.** Training pipeline and downstream deployment of CF-JEPA.

Let $\mathcal{D} = \{x^{(n)}\}_{n=1}^{N}$ denote a training set of $N$ time series, where each sample $x \in R^{T\times C}$ has length $T$ and $C$ channels. The objective is to learn an online encoder $f_{\theta_o}: R^{L\times C} \to R^{L\times d}$ parameterized by $\theta_o$, which produces a representation $z \in R^{L\times d}$ for an input subsequence of length $L$, where $L$ is determined by random cropping at training time. In addition to the online encoder, CF-JEPA maintains a target encoder $f_{\theta_t}$ with identical architecture, whose parameters $\theta_t$ are updated as the EMA of $\theta_o$ and receive no gradient. During pretraining, length-$L$ crops are encoded by the online encoder, and the resulting representations are passed through a forward predictor to predict the representations of the future portion of the same series, as encoded by the target encoder. After pretraining, the learned encoders are evaluated in three downstream tasks: classification, forecasting, and anomaly detection. Throughout this paper, bold lowercase letters are used for vectors and sequences (e.g., $x$, $z$), uppercase for sets and matrices, and $||\cdot||$ to denote the $\mathrm{L}_2$ norm unless otherwise stated.

*3.2 Crop-based view generation*

CF-JEPA generates training views through random forward cropping, without applying data augmentation. Each training sample $x \in R^{T\times C}$ is first normalized to zero mean and unit variance. From each sample, $n_{crops}$ random subsequences are extracted by uniformly sampling a crop ratio $r \in [r_{min}, r_{max}]$ and a start position s satisfying $s + \lfloor r \cdot T \rfloor \leq T$, yielding crops $x_{crop} \in R^{L\times C}$ of length $L = \lfloor r \cdot T \rfloor$. Forward cropping is designed to preserve the temporal direction of the underlying series, whereas the random ratio and start position provide a diverse set of context views.

*3.3 Forward predictive training*

In this stage, the online encoder is trained to predict future segments of the same time series in representation space, using targets provided by the EMA target encoder.

***3.3.1 Encoder Architecture.*** The encoder $f_\theta$ is built upon a stack of *D* blocks, each consisting of multi-scale dilated depthwise convolutions (DWConv), designed to capture temporal patterns at multiple granularities. The input $x_{crop}$ is first projected onto a hidden dimension $h$ via a linear layer and then passed through *D* blocks. Each block applies three parallel DWConvs with kernel sizes {3, 9, 15} to capture patterns at multiple temporal scales, followed by batch normalization, GELU activation, pointwise convolution, and residual connections. The kernel sizes {3, 9, 15} are designed to cover short-, mid-, and long-range temporal patterns within a single block, whereas depthwise-pointwise factorization preserves the representational capacity with a much smaller parameter footprint than a standard convolution. The dilation rate increases as $2^i$ at the *i*-th block to expand the receptive field with depth. A final linear layer projects the output to the representation dimension *d*, producing $z_{crop} = f_{\theta_o}(x_{crop}) \in R^{L \times d}$. The target encoder $f_{\theta_t}$ shares this architecture, and its parameters are updated after each optimization step as:

$$\theta_t \leftarrow m \cdot \theta_t + (1 - m) \cdot \theta_o \tag{1}$$

where *m* is the EMA decay rate that follows a cosine schedule from $m_{base}$ to 1 over the course of training. The target encoder receives no gradient and encodes the full time series *x* to provide stable prediction targets $z_{full} = f_{\theta_t}(x) \in R^{T \times d}$. Maintaining the target encoder as an EMA of the online encoder provides prediction targets that evolve slowly relative to the online updates, preventing the optimization from chasing rapidly shifting representations.

***3.3.2 Forward Predictive Loss.*** Given a crop encoded as $z_{crop} \in R^{L \times d}$, CF-JEPA predicts the representation of the future segment of the same time series, as encoded by the target encoder. Specifically, the portion of the time series following the crop is divided into three equal-length zones $\{Z_1, Z_2, Z_3\}$, corresponding to short-, mid-, and long-horizon futures relative to the crop. Three linear predictors $\{P_1, P_2, P_3\}$ are assigned to these zones in a one-to-one mapping, where each $P_i$ is exclusively responsible for predicting the target representation in zone $Z_i$. The three-zone partition reflects a natural division of the forecast window into short-, mid-, and long-horizon ranges, and the one-to-one mapping encourages each predictor to specialize in its assigned temporal range rather than averaging across multiple horizons. Each predictor is initialized near the identity, takes $z_{crop}$ as input and produces a prediction $\widehat{z_i} = P_i(z_{crop})$, which is matched against the target representation $z_{full}[Z_i]$ under the $L_1$ distance with both vectors normalized to unit norm:

$$\mathcal{L}_{horizon} = \frac{1}{3}\sum_{i=1}^{3} \left\| \text{normalize}\left(P_i(z_{crop})\right) - \text{normalize}\left(z_{full}[Z_i]\right) \right\|_1 \quad (2)$$

Near-identity linear initialization preserves the temporal alignment between the context and target representations, encouraging the encoder to absorb the forecasting structure rather than the predictor.

### *3.4 Regularization and training objective*

Pure prediction-based learning is prone to representational collapse [5, 28]. In this failure mode, the encoder converges to a trivial solution that minimizes the prediction loss but loses discriminative information. To prevent this, CF-JEPA uses the variance and covariance terms of VICReg [6] on the mean-pooled crop representations $\bar{z} = \frac{1}{L}\sum_t z_{crop}[t] \in R^d$. The variance regularization encourages the standard deviation of each dimension to be at least one across the batch, preventing dimension-wise collapse:

$$\mathcal{L}_{var} = \frac{1}{d}\sum_{j=1}^{d} max\left(0,1 - std(\overline{z_j})\right) \quad (3)$$

Covariance regularization decorrelates the representation dimensions by penalizing off-diagonal entries of the empirical covariance matrix $\text{Cov}(\bar{z})$:

$$\mathcal{L}_{cov} = \frac{1}{d}\sum_{i \neq j} \text{Cov}(\bar{z})_{ij}^2 \quad (4)$$

In addition, a multi-scale invariance loss is applied to encourage crops drawn from the same time series to map to similar representations across multiple temporal scales. For each pool size $p \in \{2, 4, 8\}$, crop representations are reduced via adaptive average pooling to obtain $z_{crop}^{(p)} \in R^{p\times d}$, and the variance across crops of the same sample is minimized:

$$\mathcal{L}_{inv} = \frac{1}{3} \sum_{p\in\{2,4,8\}} \mathrm{Var}_{crops}\left(z_{crop}^{(p)}\right) \tag{5}$$

where $Var_{crops}(\cdot)$ denotes the element-wise variance computed across the $n_c$ crops drawn from the same time series. The pool sizes {2, 4, 8} are chosen so that invariance is enforced simultaneously at coarse, mid, and fine temporal granularities, preventing the model from collapsing at any single scale. The overall training objective combines the prediction loss with three regularization terms:

$$\mathcal{L} = w_{horizon}(t) \cdot \mathcal{L}_{horizon} + \lambda_{var} \cdot \mathcal{L}_{var} + \lambda_{cov} \cdot \mathcal{L}_{cov} + \lambda_{inv} \cdot \mathcal{L}_{inv} \tag{6}$$

where $w_{horizon}(t) = 1 - \frac{t}{T_{max}}$ linearly anneals the horizon weight from one to zero across training epochs $t \in [0, T_{max}]$. The model was optimized using AdamW [30] under a cosine-learning-rate schedule, with gradient accumulation applied when the effective batch size was small to stabilize the VICReg statistics. The full pretraining procedure is summarized in Algorithm 1.

**Algorithm 1.** CF-JEPA Pretraining

**Input:** Unlabeled time series $\mathcal{D} = \{x^{(n)}\}$, epochs *E*, number of crops $n_c$, crop ratio $[r_{min}, r_{max}]$, zones *K*, EMA decay *m*, loss weights $(\lambda_{var}, \lambda_{cov}, \lambda_{inv})$

**Output:** Online encoder $f_{\theta_o}$, Target encoder $f_{\theta_t}$

1 Initialize online encoder $f_{\theta_o}$, target encoder $f_{\theta_t} \leftarrow f_{\theta_o}$, predictor $\{P_k\}$

2 **for** epoch = 1 to *E d*

3 $w \leftarrow 1 - \mathrm{epoch}/\mathrm{E}$ // horizon annealing

4 **for** mini-batch $x \in \mathcal{D}$ **do**

5 Sample $n_c$ random forward crops $\{x_{c,i}\}$ per sample

6 $z_{c,i} \leftarrow f_{\theta_o}(x_{c,i})$ **for** $i = 1, \ldots, n_c$

7 $z_{full} \leftarrow \text{stopgrad}\left(f_{\theta_t}(x)\right)$ // target representation

8 $\mathcal{L}_{horizon} \leftarrow 0$

9 **for** each crop $i$, zone $k$ **do**

10 $\widehat{z_{i,k}} \leftarrow P_k(z_{c,i})$

11 $z_{i,k}^{tgt} \leftarrow z_{full}[zone_k]$

12 $\mathcal{L}_{horizon} \leftarrow \mathcal{L}_{horizon} + \left\|\text{norm}(\widehat{z_{i,k}}) - \text{norm}(z_{i,k}^{tgt})\right\|_1$ // Eq. (2)

13 **end for**

14 $\mathcal{L}_{var}, \mathcal{L}_{cov} \leftarrow \text{VICReg}\left(\text{pool}(\{z_{c,i}\})\right)$ // Eq. (3), (4)

15 $\mathcal{L}_{inv} \leftarrow \text{MultiScaleInvariance}(\{z_{c,i}\}; \text{pool} \in \{2,4,8\})$ // Eq. (5)

16 $\mathcal{L}_{total} \leftarrow w \cdot \mathcal{L}_{horizon} + \lambda_{var} + \lambda_{cov} + \lambda_{inv}$ // Eq. (6)

17 $\theta_o \leftarrow \theta_o - \eta \cdot \nabla_\theta \mathcal{L}_{total}$

18 $\theta_t \leftarrow m \cdot \theta_t + (1 - m) \cdot \theta_o$ // Eq. (1) EMA update

19 **end for**

20 **end for**

21 **return** $f_{\theta_o}$, $f_{\theta_t}$

### *3.5 Dual-encoder downstream protocol*

Pretraining jointly produces two encoders with different representational properties: the online encoder $f_{\theta_o}$ retains discriminative features useful for sample-level classification, while the target encoder $f_{\theta_t}$, through EMA weight averaging, develops smoother temporal representations useful for per-timestep regression. Each downstream task is assigned to the encoder best suited to its inductive bias: the online encoder for classification and the target encoder for forecasting and anomaly detection. For classification, a time series *x* is encoded by the online encoder to produce $z = f_{\theta_o}(x) \in R^{T \times d}$. The representation is reduced via adaptive average pooling into $n_{bins}$ temporal bins, flattened into a vector of dimension $n_{bins} \cdot d$, and passed to a linear support vector machine (SVM). For forecasting, the target encoder is applied causally: at each time step $t$, it encodes a left-padded prefix of the time series to produce $z_t = f_{\theta_t}(x_{\leq t})$, and a Ridge regression model is fit to predict future values from $\{z_t\}$. For anomaly detection, each time series is encoded by the target encoder, and per-timestep anomaly scores are computed as distances (k-NN or Gaussian) from the training-distribution statistics. An anomaly is declared when the score exceeds a threshold based on the training mean and standard deviation.

## 4. Experiments

CF-JEPA was evaluated on three downstream tasks—classification, forecasting, and anomaly detection—using standard time-series benchmarks. Shared implementation details are provided in Section 4.1, with task-specific datasets, protocols, and results described in Sections 4.2–4.4. Ablation and analysis are presented in Sections 4.5.

### *4.1 Implementation details*

Eleven self-supervised time-series representation learning methods are compared (ten baselines and CF-JEPA) under a unified evaluation protocol across all three downstream tasks. The baselines are TS2Vec [8], CoST [21], TRep [20], InfoTS [10], SoftCLT [19], TimesURL [22], SimMTM [13], AutoTCL [9], FEI [27], and TS-JEPA [17]. The comparison is limited to self-supervised representation learning methods. Fully supervised forecasting models (e.g., PatchTST [24], iTransformer, TimesNet) were not included as they address a different problem setup, and train end-to-end forecasting models with labeled supervision rather than learning task-agnostic representations. Each baseline is re-run end-to-end rather than using results from the original papers, enabling controlled comparison under matched training and evaluation conditions. All methods use identical training conditions where possible. Method-specific hyperparameters (learning rate, hidden dimension, and architectural depth) followed each original study to ensure a fair comparison without per-dataset tuning. Where the encoder architecture allowed, the representation dimension was set to $d = 320$ across baselines to match CF-JEPA, eliminating capacity as a confounding factor. Three baselines whose architectures fix the representation dimension by design retain their original values: SimMTM at $d = 128$, FEI at $d = 1024$, and TS-JEPA at $d = 384$.

Unlike the iteration-based protocol used in several original papers (e.g., TS2Vec uses a fixed number of iterations), the number of epochs were fixed so that every method observed the full training set the same number of times regardless of the dataset size. This approach prevents large datasets from being only partially observed, whereas small datasets are observed hundreds of times under a fixed iteration budget. Table 2 summarizes the training protocol, with each row specifying one setting: values either follow each method's original paper or are applied uniformly across datasets when the original paper tunes per dataset.

**Table 2.** Training protocol shared across all methods.

| Setting | Value |
|---|---|
| Epochs | 100 (classification), 200 (forecasting, anomaly detection) |
| Batch size | 16 (fixed) |
| Seeds | 3 seeds {42, 43, 44}; mean ± std reported |
| Optimizer / LR | Each method original paper value |
| Hidden dim $h$ | $h = 64$ (most baselines); $h = 60$ (InfoTS);<br>layer-wise [64–256] (T-Rep); $h = 256$ (CF-JEPA) |
| Representation dim $d$ | $d = 320$ (most baselines + CF-JEPA);<br>architecture-fixed: SimMTM $d = 128$, FEI $d = 1024$, TS-JEPA $d = 384$ |
| Long-sequence handling | Sliding window with maximum train length = 2048 for $T > 2048$ |

To ensure fair comparison, no per-dataset hyperparameter tuning is performed for any method, and all method-specific hyperparameters follow those specified in each original paper. For methods whose original implementations tune hyperparameters per dataset (e.g., SoftCLT's $\tau_{inst}$, $\tau_{temp}$), the per-dataset table is replaced with a single fixed value applied uniformly across all datasets. In addition, some baselines required minor adjustments to their public reference implementations to faithfully reproduce the behaviors reported in the papers. For example, TS-JEPA's EMA momentum schedule is corrected to match the per-iteration update specified in the original paper, and CoST's training length is capped at $min(3000, T)$ to prevent FFT-window degeneration on short sequences. These adjustments address the faithfulness of reproduction rather than per-dataset tuning. SimMTM was evaluated only for classification because its reconstruction head produces series-level embeddings without the per-timestep representations required for forecasting and anomaly detection.

The CF-JEPA encoder consists of $D = 5$ multi-scale dilated DWConv blocks with hidden dimension $h = 256$ and representation dimension $d = 320$. AdamW is used with learning rate $2.25 \times 10^{-4}$ under a cosine schedule, with $n_{crops} = 4$ crops per sample, crop ratios uniformly sampled from [0.367, 0.418] (boosted to [0.6, 0.8] for short sequences with $T < 50$), and EMA decay $m_{base} = 0.983$. The loss weights ($\lambda_{var} = 0.081$, $\lambda_{cov} = 0.076$, and $\lambda_{inv} = 1.101$) and pooling resolution ($n_{bins} = 8$) are determined on a held-out development subset disjoint from the UCR and UEA test splits and are applied uniformly across all datasets.

## *4.2 Classification*

The UCR Time Series Classification Archive [31] and UEA Multivariate Time Series Classification Archive [32] were used, comprising 126 univariate and 26 multivariate datasets, respectively. Each dataset provides a predefined training/testing split, with per-dataset sample sizes ranging from 16 to several thousand and sequence lengths ranging from 24 to over 2,000.

An SVM classifier was trained on frozen representations produced by the encoder of each method, and test accuracy was reported. To prevent bias from a particular kernel or pooling choice, two pooling strategies were considered (max pooling over the temporal axis and adaptive average pooling into $n_{bins}$ = eight temporal bins). Combined with two SVM kernels (linear and RBF), this resulted in four evaluation configurations per method (M·L,

M·R, A·L, A·R; where M/A denote max-/adaptive-average pooling and L/R denote linear/RBF SVM kernel). For each method, every configuration was run uniformly across all datasets in the benchmark, and the single configuration that yielded the highest mean accuracy across the entire benchmark was selected as the representative configuration. This protocol gives each method its most favorable single setting while strictly avoiding per-dataset cherry-picking as the same configuration is applied to every dataset. The SVM regularization parameter C was selected via 5-fold cross-validation on the training split.

Table 3 lists the average classification accuracy and rank for the UCR 126 and UEA 26 benchmarks. CF-JEPA achieved the best average accuracy (0.819 for UCR, 0.717 for UEA) and the best average rank (3.61 on UCR, 3.54 on UEA) among self-supervised time-series representation learning methods, with the top-3 baselines (TS2Vec, TimesURL, and SoftCLT) within the Nemenyi critical difference (Figure 2).

**Table 3.** Classification accuracy on University of California, Riverside (UCR) 126 and University of East Anglia (UEA) 26 benchmarks.

| Method | UCR Eval | UCR Acc | UCR Avg Rank | Wilcoxon *p* (UCR) | UEA Eval | UEA Acc | UEA Avg Rank | Wilcoxon *p* (UEA) |
|---|---|---|---|---|---|---|---|---|
| Ours | A·L | **0.819 ± 0.001** | **3.61** | - | A·L | **0.717 ± 0.001** | **3.54** | - |
| Ts2Vec | M·L | <u>0.809 ± 0.001</u> | <u>4.61</u> | 0.019* | A·L | <u>0.704 ± 0.003</u> | <u>4.94</u> | 0.166 |
| CoST | M·L | 0.791 ± 0.001 | 5.42 | <0.001*** | M·L | 0.685 ± 0.007 | 5.65 | 0.009** |
| InfoTS | M·L | 0.787 ± 0.001 | 5.17 | <0.001*** | M·R | 0.682 ± 0.010 | 5.04 | 0.017* |
| AutoTCL | A·L | 0.727 ± 0.007 | 8.56 | <0.001*** | A·L | 0.650 ± 0.014 | 8.13 | 0.003** |
| SoftCLT | M·L | 0.808 ± 0.001 | 4.88 | 0.006** | A·L | 0.700 ± 0.006 | 5.29 | 0.089 |
| TRep | M·R | 0.803 ± 0.002 | 5.27 | <0.001*** | M·R | 0.695 ± 0.010 | 4.98 | 0.021* |
| TimesURL | M·L | 0.794 ± 0.003 | 4.63 | <0.001*** | A·L | 0.687 ± 0.004 | 5.44 | 0.074 |
| SimMTM | A·R | 0.748 ± 0.003 | 7.76 | <0.001*** | A·L | 0.645 ± 0.004 | 8.15 | <0.001*** |
| FEI | A·R | 0.687 ± 0.015 | 9.24 | <0.001*** | A·L | 0.529 ± 0.007 | 9.44 | <0.001*** |
| TS-JEPA | M·R | 0.755 ± 0.004 | 6.85 | <0.001*** | A·L | 0.678 ± 0.004 | 5.38 | 0.280 |

Table 3. Classification accuracy (mean ± std, 3 seeds, epoch 100). Eval = pooling·classifier (A=adaptive avg, M=maxpool; L=linear, R=RBF), best per (method, benchmark). Avg Rank = mean per-dataset rank (lower=better). Bold = best in column, underlined = 2nd best. Wilcoxon p vs. CF-JEPA: * $p<0.05$, ** $p<0.01$, *** $p<0.001$.

Three observations were made in this study. First, CF-JEPA leads by average rank with a notable margin (3.61 vs. next-best 4.61 on UCR, and 3.54 vs. 4.94 on UEA), indicating that the gains are not driven by a few outlier datasets but by consistently strong performance across the benchmark. Second, CF-JEPA's optimal evaluation configuration is consistently adaptive average pooling with a linear SVM (A·L) on both benchmarks. In contrast, baselines vary their optimal configuration across benchmarks (most prefer maxpool + linear (M·L) on UCR but switch to A·L on UEA), suggesting that CF-JEPA produces representations that are inherently more linearly separable along the temporal axis without relying on benchmark-specific pooling choices. Third, Wilcoxon signed-rank tests against CF-JEPA confirmed that the gains were statistically significant on UCR ($p < 0.05$ for all ten baselines, with eight at $p < 0.001$. On UEA, although CF-JEPA still achieved the best mean accuracy and average rank, the differences against TS2Vec ($p = 0.166$), SoftCLT ($p = 0.089$), TimesURL ($p = 0.074$), and TS-JEPA ($p = 0.280$) were not statistically significant under a smaller sample size (26 datasets), reflecting tighter performance margins on multivariate benchmarks.

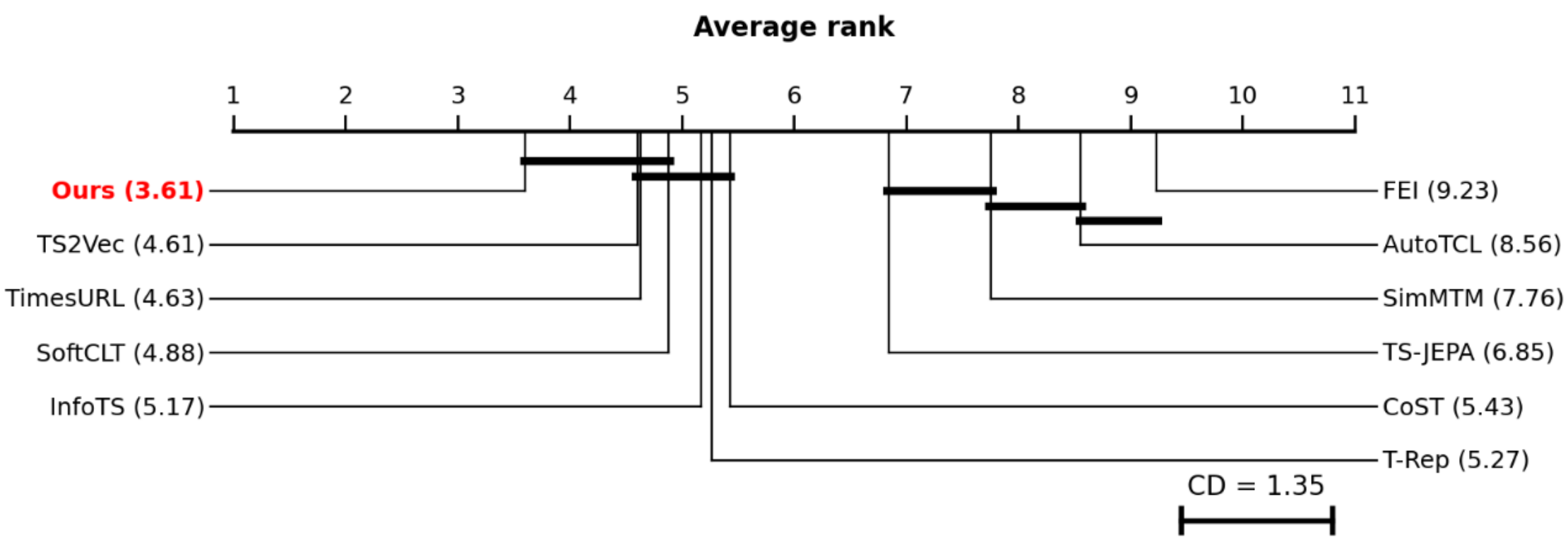

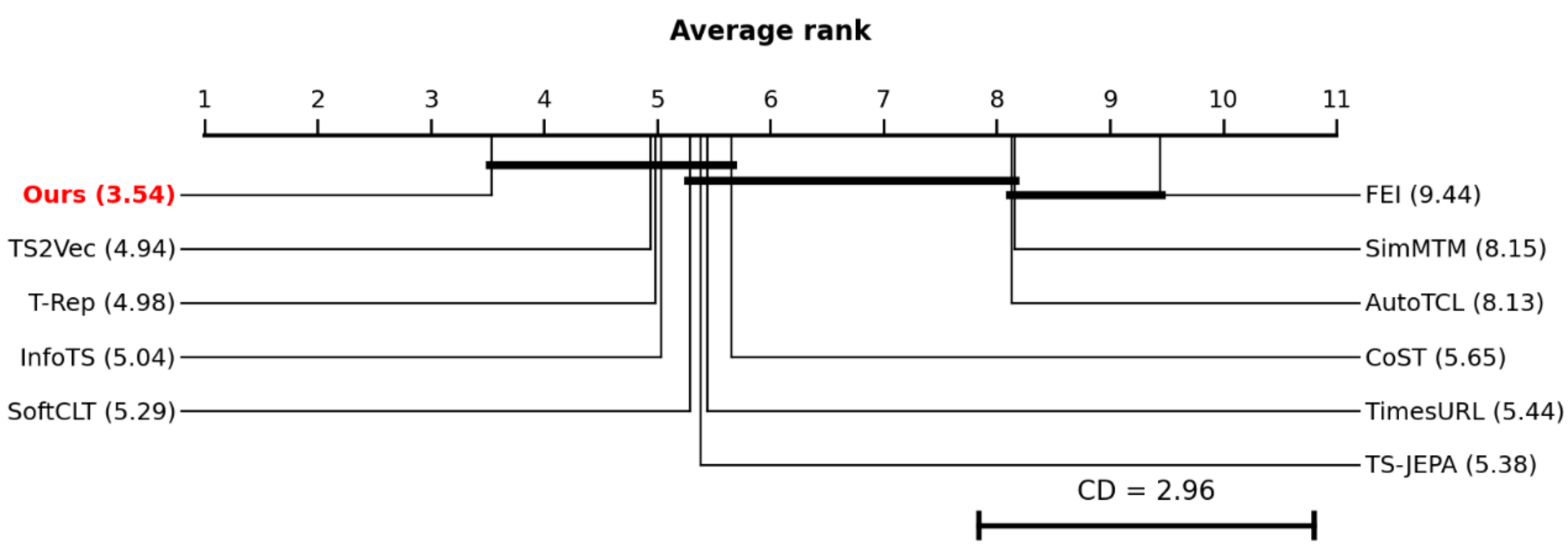

**Fig 2.** Critical difference (Nemenyi post-hoc) diagrams.

Fig. 2. (a) UCR 126 (k = 11 methods, N = 126, CD = 1.35) and (b) UEA 26 (k = 11 methods, N = 26, CD = 2.96). Lower rank indicates better performance. Horizontal bars connect methods that are statistically indistinguishable under the Nemenyi post-hoc test (alpha = 0.05). CF-JEPA (red) achieves the best average rank on both benchmarks.

**Table 4.** Classification accuracy across training budgets on UCR 126 and UEA 26.

| Method | UCR | | | UEA | | |
|---|---|---|---|---|---|---|
| | ep50 Acc | ep100 Acc | ep200 Acc | ep50 Acc | ep100 Acc | ep200 Acc |
| **Ours** | **0.816** | **0.819** | **0.813** | **0.717** | **0.717** | **0.713** |
| Ts2Vec | 0.812 | 0.809 | 0.806 | 0.700 | 0.704 | 0.710 |
| CoST | 0.792 | 0.791 | 0.792 | 0.685 | 0.685 | 0.684 |
| InfoTS | 0.787 | 0.787 | 0.785 | 0.686 | 0.682 | 0.681 |
| AutoTCL | 0.726 | 0.727 | 0.722 | 0.660 | 0.650 | 0.635 |
| SoftCLT | 0.811 | 0.808 | 0.806 | 0.693 | 0.700 | 0.707 |
| TRep | 0.807 | 0.803 | 0.803 | 0.697 | 0.695 | 0.691 |
| TimesURL | 0.794 | 0.794 | 0.797 | 0.685 | 0.687 | 0.686 |
| SimMTM | 0.746 | 0.748 | 0.751 | 0.643 | 0.645 | 0.643 |
| FEI | 0.687 | 0.687 | 0.688 | 0.525 | 0.529 | 0.527 |
| TS-JEPA | 0.752 | 0.755 | 0.757 | 0.670 | 0.678 | 0.681 |

Table 4. Each row reports mean per-dataset accuracy on UCR (126 datasets) and UEA (26 datasets) at 50, 100, and 200 epochs (3 seeds, best pooling–classifier combination per method).

Fig. 2 shows the Nemenyi post-hoc critical difference (CD) diagrams for both benchmarks. On UCR (Fig. 2(a)), CF-JEPA achieves the best average rank (3.61), with TS2Vec (4.61), TimesURL (4.63), and SoftCLT (4.88) within the critical difference (CD = 1.35), making them statistically indistinguishable from CF-JEPA under multiple-comparison correction. On UEA (Fig. 2(b)), the smaller sample size resulted in a wider CD (2.96), grouping most general-purpose methods together as statistically indistinguishable, consistent with the tighter performance margins on multivariate benchmarks. CF-JEPA was significantly separated from the bottom tiers (AutoTCL, SimMTM, and FEI) on both benchmarks.

To verify that the choice of training budget did not affect the reported ranking, all 11 methods were evaluated at 50 and 200 epochs. Ranking agreement across the three budgets is high (Kendall's $W$ = 0.97 on UCR, 0.88 on UEA; all pairwise Spearman $\rho \geq 0.76$), confirming that the reported result is not an artifact of the 100-epoch setting. CF-JEPA maintained the highest accuracy at every training budget evaluated (UCR 0.816 / 0.819 / 0.813 at epochs 50, 100, and 200; UEA 0.717 / 0.717 / 0.713), and its smallest budget accuracy matched or exceeded every baseline's best across the three budgets.

### *4.3 Forecasting*

Eight forecasting benchmarks were used, covering both univariate and multivariate long-term forecasting: ETTh1, ETTh2, ETTm1, ETTm2 [33], weather, air quality, SML, and traffic. The prediction horizons followed the dataset conventions: {24, 48, 168, 336, 720} for ETTh, {24, 48, 96, 288, 672} for ETTm, and {96, 192, 336, 720} for the remaining datasets.

For CF-JEPA, the target encoder is applied in a causal manner with left-padding (window = 200) to produce per-timestep representations $z_t = f_{\theta_t}(x_{\leq t})$. A Ridge regression model is then fit on the resulting representations to predict the next $H$ timesteps, with the ridge coefficient $\alpha$ selected on a validation split. The MSE and MAE are reported as the primary forecasting metrics.

**Table 5.** Forecasting mean squared error (MSE)/mean absolute error (MAE) on eight multivariate benchmarks.

| Method | Metric | ETTh1 | ETTh2 | ETTm1 | ETTm2 | Weather | AirQ | SML | Traffic | Avg | Rank | Rank (w/o CoST) |
|---|---|---|---|---|---|---|---|---|---|---|---|---|
| Ours | MSE | 0.78 | 1.91 | 0.55 | 0.82 | 0.29 | 1.97 | 0.64 | 0.62 | 0.95 | 4.11 | 3.17 |
| | MAE | 0.66 | 1.06 | 0.51 | 0.60 | 0.35 | 1.08 | 0.56 | 0.49 | 0.66 | 4.25 | 3.31 |
| TS2Vec | MSE | 0.80 | 1.79 | 0.60 | 0.79 | 0.29 | 1.86 | 0.63 | 0.78 | 0.94 | 5.06 | 4.11 |
| | MAE | 0.66 | 1.02 | 0.55 | 0.60 | 0.36 | 1.04 | 0.56 | 0.57 | 0.67 | 4.75 | 3.81 |
| CoST | MSE | **0.65** | **1.39** | **0.43** | **0.67** | **0.25** | **1.53** | **0.55** | **0.50** | **0.75** | **1.25** | - |
| | MAE | **0.59** | **0.91** | **0.45** | **0.54** | **0.32** | **0.92** | **0.51** | **0.41** | **0.58** | **1.28** | - |
| InfoTS | MSE | 0.81 | 1.85 | 0.59 | 0.78 | 0.30 | 1.92 | 0.71 | 0.84 | 0.97 | 6.14 | 5.19 |
| | MAE | 0.66 | 1.04 | 0.53 | 0.59 | 0.36 | 1.06 | 0.59 | 0.60 | 0.68 | 5.89 | 4.94 |
| AutoTCL | MSE | 0.87 | 2.05 | 0.64 | 1.07 | 0.33 | 2.25 | 0.68 | 0.77 | 1.08 | 7.69 | 6.69 |
| | MAE | 0.70 | 1.10 | 0.57 | 0.71 | 0.38 | 1.15 | 0.57 | 0.57 | 0.72 | 7.78 | 6.78 |
| SoftCLT | MSE | 0.78 | 1.75 | 0.60 | 0.79 | 0.29 | 1.85 | 0.62 | 0.78 | 0.93 | 4.44 | 3.50 |
| | MAE | 0.65 | 1.02 | 0.55 | 0.60 | 0.35 | 1.04 | 0.55 | 0.57 | 0.67 | 4.33 | 3.39 |
| TRep | MSE | 0.78 | 1.89 | 0.57 | 0.75 | 0.28 | 1.75 | 0.60 | 0.61 | 0.90 | 3.22 | **2.22** |
| | MAE | 0.66 | 1.06 | 0.52 | 0.59 | 0.35 | 1.00 | 0.54 | 0.48 | 0.65 | 3.67 | **2.67** |
| TimesURL | MSE | 0.79 | 1.83 | 0.59 | 0.80 | 0.30 | 1.97 | 0.66 | 0.78 | 0.96 | 5.64 | 4.67 |
| | MAE | 0.65 | 1.02 | 0.53 | 0.59 | 0.36 | 1.08 | 0.57 | 0.57 | 0.67 | 5.14 | 4.19 |
| FEI | MSE | 1.03 | 4.62 | 0.94 | 2.73 | 0.37 | 3.97 | 0.92 | 0.76 | 1.92 | 8.83 | 7.83 |
| | MAE | 0.79 | 1.74 | 0.74 | 1.28 | 0.42 | 1.55 | 0.69 | 0.57 | 0.97 | 9.22 | 8.22 |
| TS-JEPA | MSE | 1.16 | 2.40 | 0.86 | 1.41 | 0.44 | 1.93 | 1.04 | 1.28 | 1.31 | 8.61 | 7.61 |
| | MAE | 0.86 | 1.27 | 0.68 | 0.88 | 0.45 | 1.06 | 0.77 | 0.79 | 0.85 | 8.69 | 7.69 |

Table 5. Forecasting mean squared error (MSE)/mean absolute error (MAE) on eight multivariate benchmarks. Mean over 3 seeds at 200 epochs. Avg = mean across the 8 datasets. Rank = mean per-(dataset, prediction-length) rank (lower = better); Rank (w/o CoST) excludes CoST from ranking. Bold = best in column, underlined = 2nd best (within MSE rows / MAE rows separately). SimMTM is excluded as it does not produce per-timestep representations.

**Table 6.** Forecasting MSE/MAE on eight univariate benchmarks.

| Method | Metric | ETTh1 | ETTh2 | ETTm1 | ETTm2 | Weather | AirQ | SML | Traffic | Avg | Rank | Rank (w/o CoST) |
|---|---|---|---|---|---|---|---|---|---|---|---|---|
| Ours | MSE | 0.16 | **0.17** | 0.13 | **0.11** | **0.01** | 0.81 | **1.08** | 0.40 | 0.36 | 3.17 | **2.56** |
| | MAE | 0.32 | **0.32** | 0.26 | **0.23** | 0.07 | 0.62 | 0.79 | 0.41 | 0.38 | 3.19 | **2.58** |
| TS2Vec | MSE | 0.17 | 0.19 | 0.15 | 0.13 | **0.01** | 0.88 | 1.17 | 0.41 | 0.39 | 5.60 | 4.75 |
| | MAE | 0.32 | 0.33 | 0.29 | 0.25 | **0.06** | 0.65 | 0.84 | 0.42 | 0.40 | 5.50 | 4.64 |
| CoST | MSE | **0.12** | 0.18 | **0.10** | **0.11** | **0.01** | **0.74** | **1.08** | **0.33** | **0.33** | **2.36** | - |
| | MAE | **0.26** | **0.32** | **0.22** | **0.23** | 0.07 | **0.61** | **0.78** | **0.35** | **0.36** | **2.25** | - |
| InfoTS | MSE | 0.17 | 0.19 | 0.14 | 0.12 | **0.01** | 0.87 | **1.08** | 0.44 | 0.38 | 4.78 | 3.94 |
| | MAE | 0.33 | 0.33 | 0.26 | 0.24 | **0.06** | 0.65 | **0.78** | 0.45 | 0.39 | 4.64 | 3.81 |
| AutoTCL | MSE | 0.96 | 0.20 | 0.17 | 0.20 | **0.01** | 1.02 | 1.34 | 0.50 | 0.55 | 7.72 | 6.89 |
| | MAE | 0.57 | 0.35 | 0.29 | 0.32 | **0.06** | 0.72 | 0.86 | 0.49 | 0.46 | 7.56 | 6.69 |
| SoftCLT | MSE | 0.18 | 0.20 | 0.15 | 0.13 | **0.01** | 0.87 | 1.31 | 0.40 | 0.41 | 6.01 | 5.12 |
| | MAE | 0.34 | 0.34 | 0.29 | 0.25 | **0.06** | 0.65 | 0.88 | 0.42 | 0.40 | 5.96 | 5.10 |
| TRep | MSE | 0.18 | 0.20 | 0.14 | 0.14 | **0.01** | 0.85 | 1.17 | 0.37 | 0.38 | 5.86 | 4.89 |
| | MAE | 0.34 | 0.35 | 0.27 | 0.26 | 0.07 | 0.64 | 0.85 | 0.39 | 0.40 | 5.94 | 4.94 |
| TimesURL | MSE | 0.15 | 0.18 | 0.12 | 0.12 | **0.01** | 0.86 | 1.15 | 0.44 | 0.38 | 3.85 | 3.01 |
| | MAE | 0.30 | 0.33 | 0.25 | 0.24 | **0.06** | 0.65 | 0.81 | 0.44 | 0.38 | 3.79 | 2.93 |
| FEI | MSE | 0.26 | 0.30 | 0.13 | 0.13 | **0.01** | 0.91 | 6.77 | 0.47 | 1.12 | 7.76 | 6.86 |
| | MAE | 0.41 | 0.42 | 0.26 | 0.26 | 0.07 | 0.68 | 1.56 | 0.50 | 0.52 | 7.86 | 6.92 |
| TS-JEPA | MSE | 0.26 | 0.25 | 0.13 | 0.18 | **0.01** | 0.99 | 1.81 | 0.68 | 0.54 | 7.89 | 6.97 |
| | MAE | 0.44 | 0.40 | 0.28 | 0.33 | **0.06** | 0.71 | 1.13 | 0.58 | 0.49 | 8.31 | 7.39 |

Table 6. Forecasting MSE/MAE on eight univariate benchmarks. Same protocol and column conventions as Table 5.

**Table 7.** Forecasting MSE across training budgets

| Method | Multivariate | | | Univariate | | |
|---|---|---|---|---|---|---|
| | ep50 MSE | ep200 MSE | ep400 MSE | ep50 MSE | ep200 MSE | ep400 MSE |
| **Ours** | 0.942 | 0.949 | 0.957 | 0.356 | 0.359 | 0.360 |
| Ts2Vec | 0.977 | 0.944 | 0.932 | 0.372 | 0.388 | 0.371 |
| CoST | **0.739** | **0.746** | **0.748** | **0.320** | **0.333** | **0.334** |
| InfoTS | 0.973 | 0.975 | 0.980 | 0.374 | 0.375 | 0.379 |
| AutoTCL | 1.044 | 1.083 | 1.033 | 0.439 | 0.550 | 0.876 |
| SoftCLT | 0.963 | 0.932 | 0.927 | 0.373 | 0.406 | 0.373 |
| TRep | 0.895 | 0.905 | 0.926 | 0.360 | 0.383 | 0.376 |
| TimesURL | 0.959 | 0.965 | 0.968 | 0.371 | 0.376 | 0.376 |
| FEI | 1.901 | 1.918 | 1.918 | 0.776 | 1.122 | 1.122 |
| TS-JEPA | 1.162 | 1.315 | 1.344 | 0.505 | 0.537 | 0.531 |

Table 7. Mean MSE at 50, 200, and 400 epochs (3 seeds; mean across the eight datasets, with each dataset entry averaged over prediction horizons). SimMTM is excluded as it does not produce per-timestep representations.

Tables 5 and 6 present the forecasting MSE for the eight benchmark datasets under multivariate and univariate settings, respectively. On the multivariate task (Table 5), CF-JEPA achieves an average MSE of 0.95 and an average rank of 4.11, placing third among the ten methods. CoST leads the benchmark with an Avg Rank of 1.25 (MSE 0.75), followed by TRep with an Avg Rank of 3.22. CoST's forecasting-specific design (frequency-domain seasonal-trend disentanglement) accounts for its lead. The multivariate ETTh2 gap is particularly pronounced (CoST 1.39 vs. CF-JEPA 1.91), a known field result reflecting the seasonal-trend regularity that CoST's frequency-domain decomposition is designed to exploit.

On the univariate task (Table 6), CF-JEPA achieves an average MSE of 0.36 and an average rank of 3.17, placing second overall behind CoST (MSE 0.33, Rank 2.36). CoST's forecasting-specific design accounts for its lead. Among the remaining general-purpose self-supervised methods, CF-JEPA ranks first with an Avg Rank of 2.56 (w/o CoST), ahead of TimesURL (3.01) and InfoTS (3.94), showing that forward-predictive representations transfer effectively to univariate long-horizon forecasting.

To verify that the choice of training budget did not affect the reported ranking, all methods were evaluated at 50 and 400 epochs (Table 7). Ranking agreement across the three budgets is high (Kendall's W = 0.95 multivariate, 0.94 univariate; all pairwise Spearman $\rho \geq 0.85$), confirming that the reported result is not an artifact of the 200-epoch setting.

*4.4 Anomaly detection*

KPI [34] and Yahoo Webscope S5 [35] were used with delay tolerances of seven and three time steps, respectively, following prior work. Per-timestep representations $z_t \in R^D$ are extracted from each series, and two unsupervised anomaly scores are evaluated:

k-NN distance ($k$ = 5): the L2 distance to the $k$-th nearest training time step after $z$-normalization, with training representations subsampled to 50,000 timesteps.

Gaussian distance: the standardized L2 distance to the mean of training representations.

An anomaly is declared when , following [8, 36], with consecutive-detection suppression within the delay window. Delay-adjusted F1, along with precision and recall, was reported as the primary metrics.$s_t > \mu_{train} + 4 \cdot \sigma_{train}$

**Table 8.** Anomaly detection F1, precision and recall on KPI and Yahoo benchmarks.

| Method | Scoring | KPI F1 | KPI P | KPI R | Yahoo F1 | Yahoo P | Yahoo R | Avg F1 | F1 Rank |
|---|---|---|---|---|---|---|---|---|---|
| Ours | k-NN | 0.685 | **0.826** | 0.585 | **0.317** | **0.207** | 0.680 | **0.501** | **1.50** |
| | Gaussian | 0.541 | 0.812 | 0.407 | 0.378 | 0.260 | 0.694 | 0.459 | 2.00 |
| TS2Vec | k-NN | 0.613 | 0.797 | 0.499 | 0.251 | 0.150 | **0.783** | 0.432 | 3.50 |
| | Gaussian | 0.356 | 0.785 | 0.232 | 0.314 | 0.203 | 0.704 | 0.335 | 4.50 |
| CoST | k-NN | 0.336 | 0.215 | 0.764 | 0.089 | 0.048 | 0.633 | 0.213 | 8.00 |
| | Gaussian | 0.135 | 0.217 | 0.098 | 0.091 | 0.049 | 0.598 | 0.113 | 8.50 |
| InfoTS | k-NN | 0.525 | 0.402 | 0.780 | 0.171 | 0.099 | 0.635 | 0.348 | 5.00 |
| | Gaussian | 0.350 | 0.700 | 0.235 | 0.330 | 0.230 | 0.593 | 0.340 | 4.50 |
| AutoTCL | k-NN | 0.404 | 0.453 | 0.367 | 0.081 | 0.046 | 0.392 | 0.242 | 8.00 |
| | Gaussian | 0.119 | 0.284 | 0.076 | 0.153 | 0.111 | 0.286 | 0.136 | 8.00 |
| SoftCLT | k-NN | 0.593 | 0.501 | 0.727 | 0.093 | 0.050 | 0.686 | 0.343 | 5.50 |
| | Gaussian | 0.406 | **0.870** | 0.266 | 0.188 | 0.112 | 0.597 | 0.297 | 5.00 |
| T-Rep | k-NN | 0.627 | 0.571 | 0.713 | 0.106 | 0.058 | 0.650 | 0.367 | 4.00 |
| | Gaussian | 0.514 | 0.801 | 0.382 | 0.211 | 0.136 | 0.484 | 0.363 | 4.00 |
| TimesURL | k-NN | **0.692** | 0.712 | 0.673 | 0.286 | 0.178 | 0.718 | 0.489 | **1.50** |
| | Gaussian | **0.641** | 0.811 | **0.530** | **0.401** | **0.280** | **0.713** | **0.521** | **1.00** |
| FEI | k-NN | 0.504 | 0.369 | 0.792 | 0.049 | 0.026 | 0.666 | 0.277 | 8.00 |
| | Gaussian | 0.153 | 0.530 | 0.090 | 0.150 | 0.088 | 0.535 | 0.152 | 7.50 |
| TS-JEPA | k-NN | 0.331 | 0.200 | **0.970** | 0.044 | 0.023 | 0.649 | 0.188 | 10.00 |
| | Gaussian | 0.067 | 0.567 | 0.036 | 0.016 | 0.009 | 0.053 | 0.041 | 10.00 |

Table 8. Delay-adjusted F1, precision, and recall on KPI (delay tolerance 7) and Yahoo Webscope S5 (delay tolerance 3), mean over 3 seeds at 200 epochs. Two scoring functions are evaluated: k-NN distance (k = 5) and Gaussian distance. Avg F1 = mean across the two benchmarks. Avg F1 Rank = mean rank across benchmarks (lower = better). Bold indicates the best result in the column, underlined indicates second best. SimMTM is excluded as it does not produce per-timestep representations.

Table 8 presents anomaly detection F1, precision, and recall on the KPI and Yahoo benchmarks using two scoring functions (k-NN distance and Gaussian distance). CF-JEPA is tied for the best Avg F1 Rank under k-NN scoring (Rank 1.50, Avg F1 = 0.501) and ranks second under Gaussian scoring (Rank 2.00, Avg F1 = 0.459) among ten methods. Per-benchmark performance is strong on both datasets. On KPI, CF-JEPA achieves Avg F1 = 0.685 under k-NN scoring, essentially tied with TimesURL (0.692) for the lead, with both methods substantially ahead of the next tier (T-Rep 0.627, TS2Vec 0.613, SoftCLT 0.593). Under Gaussian scoring on KPI, CF-JEPA ranks second (0.541) behind TimesURL (0.641). On Yahoo Webscope S5, CF-JEPA leads under k-NN scoring (F1 = 0.317 vs. TimesURL 0.286, surpassing all baselines) and ranks second under Gaussian scoring (F1 = 0.378 vs. TimesURL 0.401). The anomaly-detection leader depends on the scoring function. CF-JEPA leads under k-NN scoring (Avg F1 0.501 vs. TimesURL 0.489), while TimesURL leads under Gaussian scoring (0.521 vs. CF-JEPA 0.459). This asymmetry reflects complementary inductive biases: CF-JEPA's forward-predictive representations interact favorably with local-neighborhood scoring (k-NN), while TimesURL's reconstruction-based representations align better with parametric density scoring (Gaussian). These two frameworks emerge as the strongest general-purpose self-supervised methods for time-series anomaly detection, with no notable margin between them when each is paired with its preferred scoring function. Notably, CoST, the leader in forecasting, exhibits the weakest performance on anomaly detection (Avg F1 Rank 8.00 k-NN, 8.50 Gaussian), indicating that forecasting-specialized inductive biases (frequency-domain seasonal-trend disentanglement) do not transfer to anomaly detection. In contrast, general-purpose self-supervised representations (CF-JEPA, TimesURL, SoftCLT, TS2Vec) generalize more reliably across all three downstream tasks.

### *4.5 Ablation study*

To validate the design choices of CF-JEPA, five ablations were conducted along complementary axes: (i) loss-component removal, (ii) predictor design, (iii) predictor capacity, (iv) encoder backbone, and (v) online versus target encoder utilization. All ablations followed the same training and evaluation protocols as the main experiments (three seeds, 100 epochs for classification, and 200 epochs for forecasting). Classification was reported on UCR 126 and UEA 26 (best pooling/classifier per benchmark) and forecasting on the four ETT benchmarks (MSE averaged across prediction horizons).

### *4.5.1 Loss component ablation*

Table 9 reports the impact of removing regularization losses from the full CF-JEPA model. Each row removes one loss term (multi-scale invariance or VICReg [6] variance/covariance) while keeping the encoder and JEPA-defining components (forward predictor and EMA target) intact, thus isolating the marginal contribution of each loss to both classification and forecasting. The predictor design alternatives are presented in Table 10, and the role of the EMA target is quantified in Table 13.

**Table 9.** CF-JEPA loss component ablation.

| Variant | Cls UCR (Acc) | Cls UEA (Acc) | Fcst Multi MSE (ETT4) | Fcst Uni MSE (ETT4) | Δ vs. Full |
|---|---|---|---|---|---|
| Full (CF-JEPA) | **0.819** | **0.717** | **1.016** | **0.145** | - |
| w/o Invariance loss | 0.81 | 0.706 | 1.023 | 0.146 | Cls −0.8/−1.1pp; Fcst +0.7/+0.4% |
| w/o VICReg | 0.799 | 0.698 | 1.057 | 0.146 | Cls −2.0/−1.9pp; Fcst +4.0/+0.5% |

Table 9. Effect of removing each regularization loss from the full CF-JEPA model (3 seeds; 100 epochs for classification, 200 epochs for forecasting). Each row removes one loss term while keeping the encoder and the JEPA-defining components (forward predictor, EMA target) intact. Classification reports mean accuracy on UCR 126 and UEA 26; forecasting reports MSE averaged across prediction horizons on the four ETT benchmarks. Bold indicates the best result in each column.

Two patterns emerge. First, the VICReg variance/covariance is the single dominant regularizer for classification. Its removal drops UCR accuracy by 2.0 percentage points (0.819 → 0.799) and UEA accuracy by 1.9 percentage points (0.717 → 0.698), the largest classification penalty among the loss ablations. The multi-scale invariance loss provides a more modest but consistent benefit, with its removal decreasing UCR by 0.8 percentage points (0.819 → 0.810) and UEA by 1.1 percentage points (0.717 → 0.706). Together, these two losses prevent representation collapse and underpin the linear separability that supports the SVM classification head. Second, forecasting is largely insensitive to changes in loss components. Across both regularizer removals, the multivariate ETT MSE remained within 4.0% of the full configuration (1.023 and 1.057 vs. 1.016, respectively), and the univariate MSE remained essentially unchanged (0.145–0.146). This contrasts sharply with the 27% multivariate MSE gap between online and target encoders reported in Section 4.5.5 (Table 13): the EMA target, not the regularization losses, is the component that meaningfully differentiates forecasting performance among trained encoders.

This finding directly motivates the asymmetric utilization protocol introduced in Section 3.5 and quantified in Section 4.5.5.

*4.5.2 Predictor design*

Table 10 compares CF-JEPA's three-zone forward latent predictor with three alternative predictor paradigms: the single-zone forward variant, masked-latent prediction (TS-JEPA style), and masked-input reconstruction (TimeMAE style). This ablation isolates the choice of the predictive objective from the rest of the framework with the encoder backbone, regularization, and training schedule held fixed, thus providing direct evidence of whether forward prediction outperforms masking-based alternatives within the JEPA framework.

Three findings are noteworthy. First, within our training harness (with shared encoder, regularization, and schedule), the 3-zone forward predictor outperforms the masked-latent variant by 0.819 vs. 0.789 on UCR (−3.0 percentage points) and 0.717 vs. 0.706 on UEA (−1.1 percentage points), suggesting that the forward direction is a more compatible inductive bias for time-series representation learning within this framework. Second, multi-zone (3-zone) forward prediction primarily benefits forecasting. The single-zone variant achieves essentially equivalent classification accuracy (−0.3 percentage points UCR, −0.2 percentage points UEA) while incurring a clear forecasting penalty (1.050 vs. 1.016, +3.3% MSE on multivariate ETT). This indicates that multi-horizon supervision provides a richer training signal for forecasting deployment, whereas a single zone is sufficient for classification deployment. Third, masked-input reconstruction (TimeMAE style) underperforms forward prediction on classification (−1.9 percentage points UCR, −0.5 percentage points UEA) and univariate forecasting (+1.2% MSE), and shows essentially no difference on multivariate forecasting MSE (1.015 vs. 1.016, within 0.1%). This suggests that input-space reconstruction does not transfer the multivariate forecasting advantage of its source design to our framework.

**Table 10.** Predictor design ablation.

| Variant | Cls UCR (Acc) | Cls UEA (Acc) | Fcst Multi MSE (ETT4) | Fcst Uni MSE (ETT4) | $\Delta$ vs. Full |
|---|---|---|---|---|---|
| 3-zone forward (CF-JEPA) | **0.819** | **0.717** | 1.016 | **0.145** | - |
| 1-zone forward | 0.815 | 0.715 | 1.05 | 0.148 | Cls −0.3/−0.2pp; Fcst +3.3/+1.8% |
| Masked latent (TS-JEPA style) | 0.789 | 0.706 | 1.075 | 0.151 | Cls −3.0/−1.1pp; Fcst +5.8/+3.9% |
| Masked input (TimeMAE style) | 0.799 | 0.712 | **1.015** | 0.147 | Cls −1.9/−0.5pp; Fcst −0.1/+1.2% |

Table 10. Comparison of CF-JEPA's 3-zone forward latent predictor with three alternatives: single-zone forward, masked-latent prediction (TS-JEPA style), and masked-input reconstruction (TimeMAE style). The encoder backbone, regularization weights, and training schedule are held constant. Classification reports mean accuracy on UCR 126 and UEA 26; forecasting reports MSE on ETT. Bold indicates the best result in each column.

Note that the masked-latent and masked-input variants are reimplemented within the CF-JEPA harness; their hyperparameters are not separately optimized for their respective objectives. This comparison should be interpreted as a design-choice evaluation within our framework, rather than a direct comparison with TS-JEPA or TimeMAE, which are published systems.

Across the four metrics, the 3-zone forward variant achieved either the best or near-best results for every measure, demonstrating that forward prediction with multi-horizon supervision is the most balanced predictor design within the JEPA framework for time series.

### *4.5.3 Predictor capacity*

Table 11 compares CF-JEPA's linear near-identity predictor with two capacity-scaled alternatives: a 2-layer MLP (Linear → GELU → Linear) and a 1-layer Transformer encoder ($n_{head}$= 4). The encoder backbone, regularization, and training schedule were fixed to isolate the effect of predictor capacity. This comparison examines whether higher-capacity predictors, as used in image and video JEPA, improve or degrade time-series representation learning within the framework.

Two findings emerge. First, the lightweight linear predictor matched or slightly outperformed both higher-capacity alternatives across all four metrics within our tested capacity range: Cls UCR (0.819 vs. 0.816 / 0.817), Cls UEA (0.717 vs. 0.715 / 0.716), Fcst Multi MSE (1.016 vs. 1.020 / 1.017), and Fcst Uni MSE (0.145 vs. 0.148 / 0.147). Second, the differences are small (Cls within 0.3 percentage points, Fcst MSE +0.1–1.9%), and no measurable benefit is observed from the added capacity under our training schedule. This is consistent with the JEPA convention of using lightweight predictors and contrasts with the deep predictors used in image and video JEPA (e.g., I-JEPA and V-JEPA).

**Table 11.** Predictor capacity ablation.

| Variant | Cls UCR (Acc) | Cls UEA (Acc) | Fcst Multi MSE (ETT4) | Fcst Uni MSE (ETT4) | $\Delta$ vs. Full |
|---|---|---|---|---|---|
| Linear (CF-JEPA) | **0.819** | **0.717** | **1.016** | **0.145** | - |
| 2-layer MLP | 0.816 | 0.715 | 1.02 | 0.148 | Cls −0.2/−0.2pp; Fcst +0.4/+1.9% |
| 1-layer Transformer | 0.817 | 0.716 | 1.017 | 0.147 | Cls −0.1/−0.1pp; Fcst +0.1/+1.5% |

Table 11. Comparison of CF-JEPA's linear near-identity predictor with two capacity-scaled alternatives: a 2-layer MLP (Linear → GELU → Linear) and a 1-layer Transformer encoder ($n_{head}$= 4). The encoder backbone, regularization weights, and training schedule are fixed. Classification reports mean accuracy on UCR 126 and UEA 26; forecasting reports MSE averaged across prediction horizons on the four ETT benchmarks. Bold indicates the best result in each column.

This is consistent with the JEPA convention of using lightweight predictors, and contrasts with the deep predictors used in image and video JEPA (e.g., I-JEPA and V-JEPA).

### *4.5.4 Encoder backbone*

Table 12 compares CF-JEPA's MultiScale dilated DWConv encoder with three alternatives: single-scale DWConv, Transformer, and TS2Vec dilated convolution encoders. The predictor (3-zone forward) and all loss weights are fixed. Unlike the predictor ablation in Table 11, which compares thin predictors following the JEPA convention, this table holds encoder capacity constant; all four backbones use hidden dimensions of 256 and a depth of 5 (matching CF-JEPA's DWConv configuration). This controlled capacity setup isolates the contribution of architectural inductive bias from the raw parameter count.

**Table 12.** Encoder backbone ablation.

| Variant | Cls UCR (Acc) | Cls UEA (Acc) | Fcst Multi MSE (ETT4) | Fcst Uni MSE (ETT4) | $\Delta$ vs. Full |
|---|---|---|---|---|---|
| MultiScale DWConv (CF-JEPA) | **0.819** | **0.717** | 1.016 | **0.145** | - |
| Single-scale DWConv | 0.817 | 0.715 | **0.976** | 0.153 | Cls −0.1/−0.2pp; Fcst −4.0/+5.2% |
| Transformer | 0.758 | 0.704 | 1.17 | 0.155 | Cls −6.1/−1.3pp; Fcst +15.1/+6.6% |
| TS2Vec encoder (Dilated Conv) | 0.756 | 0.685 | 1.116 | 0.149 | Cls −6.3/−3.2pp; Fcst +9.9/+3.1% |

Table 12. Comparison of CF-JEPA's MultiScale dilated DWConv encoder with three alternative backbones under capacity-controlled settings (hidden = 256, depth = 5, matching CF-JEPA): single-scale DWConv, Transformer ($n_{head}$= 4), and the TS2Vec dilated convolution encoder. The predictor (3-zone forward) and loss weights are held fixed. Bold indicates the best result in each column.

Three findings emerge. First, under matched capacity, DWConv-based encoders substantially outperform alternative architectures in classification. Both multi- and single-scale DWConv variants exceed the Transformer (0.819 vs. 0.758, −6.1 percentage points UCR) and the TS2Vec dilated convolution encoder (0.819 vs. 0.756, −6.3 percentage points UCR). The DWConv inductive bias and local receptive fields with channel-wise filtering appear better suited to time-series representation learning than attention-based or globally strided alternatives within our forward-prediction framework. Second, the multi-scale extension is a refinement rather than a critical contribution. MultiScale DWConv produces a

+0.2 percentage point classification gain over single-scale DWConv (within the noise margin) and is marginally worse in multivariate forecasting (1.016 vs. 0.976, +4.0% MSE). Thus, the multi-scale variant is considered a minor architectural refinement that aggregates kernel-level multi-resolution patterns The gains attributed to our backbone arise mainly from the DWConv design itself, not from the multi-scale extension. Third, even with matched capacity (hidden = 256, depth = 5), the TS2Vec dilated convolution encoder transfers poorly to our framework (−6.3 percentage points UCR, −3.2 percentage points UEA). Because hidden dimensions and depth confounds are controlled here, this gap indicates a backbone–objective compatibility issue: the TS2Vec encoder architecture, designed for the contrastive objective, does not transfer effectively to the forward-prediction objective of CF-JEPA.

This observation suggests that backbone–objective compatibility, not just the architectural family, may influence downstream performance.

#### *4.5.5 Online vs. target encoder*

Motivated by the design choice in Section 3.5, CF-JEPA's downstream protocol routes classification to the online encoder and forecasting (and anomaly detection) to the EMA target encoder. Although the two encoders share an identical architecture and are jointly produced from a single training run, their update rules differ: gradient descent for the online encoder, and exponential moving average for the target encoder. Whether this seemingly minor difference produces representations that diverge meaningfully across downstream tasks, or whether the two encoders act as effectively redundant copies, is an empirical question addressed in this section. Table 13 quantifies asymmetry by evaluating the same trained CF-JEPA model with both encoders for each downstream task. A substantial gap in opposing directions across tasks justifies task-specific routing as a non-trivial design contribution, whereas near-identical performance reduces routing to an arbitrary choice.

**Table 13.** Online vs. target encoder for downstream evaluation.

| Variant | Cls UCR (Acc) | Cls UEA (Acc) | Fcst Multi MSE (ETT4) | Fcst Uni MSE (ETT4) | Anomaly F1 k-NN | Anomaly F1 Gauss |
|---|---|---|---|---|---|---|
| Online encoder | **0.819±0.001** | **0.717±0.001** | 1.400±0.156 | 0.153±0.012 | 0.389±0.010 | 0.398±0.038 |
| Target encoder | 0.780±0.001 | 0.689±0.005 | **1.016±0.017** | **0.145±0.003** | **0.501±0.006** | **0.459±0.033** |

Table 13. The same trained CF-JEPA model is evaluated with both online and EMA target encoders on each downstream task (classification: UCR/UEA Acc; forecasting: ETT MSE; anomaly detection: F1). The opposing-direction gaps across tasks justify task-specific encoder routing as a non-trivial contribution. Bold indicates the preferred encoder for each task.

The two encoders are differentially suited to different downstream tasks in opposing directions. The online encoder achieves +3.9 percentage points higher classification accuracy on UCR (0.819 ± 0.001 vs. 0.780 ± 0.001) and +2.8 percentage points higher on UEA (0.717 ± 0.001 vs. 0.689 ± 0.005) than the target encoder. Conversely, the target encoder achieves a 27% lower multivariate forecasting MSE (1.016 ± 0.017 vs. 1.400 ± 0.156), a 5% lower univariate forecasting MSE (0.145 ± 0.003 vs. 0.153 ± 0.012), a 29% higher k-NN-scored anomaly F1 (0.501 ± 0.006 vs. 0.389 ± 0.010), and a 15% higher Gaussian-scored anomaly F1 (0.459 ± 0.033 vs. 0.398 ± 0.038) than the online encoder. The 27% reduction in multivariate forecasting MSE using the target encoder was the largest single observation in our ablation study, and this pattern extended consistently across all per-timestep deployment tasks (forecasting and anomaly detection). This pattern validates the asymmetric utilization protocol. The diagonal partitioning reflects an operational distinction between sample-level and per-timestep decision making: classification operates on a single pooled representation per series, whereas forecasting and anomaly detection operate on the per-timestep representation trajectory $z_t$— so that the geometric requirements differ accordingly (high rank for sample-level discrimination; smooth trajectories for per-timestep scoring). Through EMA weight averaging, the target encoder develops representations with smoother temporal dynamics and a lower effective rank, which benefits per-timestep regression in forecasting and anomaly detection, but comes at the cost of discriminability for sample-level classification. In contrast, the online encoder retains higher per-bin variance and effective rank, supporting the linear separability assumption underlying the SVM head used for classification.

To probe the encoder difference behind this diagonal pattern, we measured two geometric properties of the test-set representations (effective rank and adjacent-timestep cosine similarity) on a 10-dataset subset spanning the three downstream tasks (Table 14). For anomalous datasets, per-series length-weighted averaging was used to handle variable-length sequences.

**Table 14.** Online vs. target encoder representational diagnostics.

| Dataset | Type | Length | Samples | Online encoder | | | Target encoder | | | Ratio |
|---|---|---|---|---|---|---|---|---|---|---|
| | | | | $r_{\text{eff}}$ | $\overline{cos_t}$ | $\overline{cos}_{\text{bin}}$ | $r_{\text{eff}}$ | $\overline{cos_t}$ | $\overline{cos}_{\text{bin}}$ | |
| CinCECGTorso | UCR | 1639 | 1380 | 93 | 0.982 | 0.983 | 3 | 0.999 | 0.949 | 31 × |
| ECG200 | UCR | 96 | 100 | 94 | 0.945 | 0.963 | 19 | 0.996 | 0.982 | 4.9 × |
| GunPoint | UCR | 150 | 150 | 73 | 0.962 | 0.959 | 8 | 0.999 | 0.975 | 9.1 × |
| AtrialFibrillation | UEA | 640 | 15 | 85 | 0.896 | 0.972 | 7 | 0.968 | 0.957 | 12.1 × |
| SpokenArabicDigits | UEA | 93 | 2199 | 8 | 0.794 | 0.807 | 9 | 0.999 | 0.999 | 0.9 × |
| ETTh1 | Fcst | 200 | 14 | 109 | 0.703 | 0.91 | 6 | 0.936 | 0.943 | 18.2 × |
| ETTh2 | Fcst | 200 | 14 | 118 | 0.6 | 0.767 | 9 | 0.981 | 0.841 | 13.1 × |
| Weather | Fcst | 200 | 52 | 104 | 0.778 | 0.898 | 6 | 0.97 | 0.941 | 17.3 × |
| KPI | Anom. | 2961468 | 58 | 25 | 0.979 | 0.998 | 13 | 0.999 | 0.999 | 1.9 × |
| Yahoo | Anom. | 286559 | 367 | 40 | 0.96 | 0.994 | 23 | 0.989 | 0.998 | 1.7 × |
| Mean | | | | 74.9 | 0.86 | 0.925 | 10.3 | 0.984 | 0.959 | 11 × |

Table 14. Effective rank (PCA cumulative variance ≥ 90%) and adjacent-timestep cosine similarity (raw and bin-level, $n_{bins}$ = 8) of online and target encoders, measured on 10 datasets across three downstream tasks. CF-JEPA was trained with the standard schedule (epoch 100/200 for cls/fcst; single seed = 42. Rank ratio = $online_{rank}$ / $target_{rank}$ (higher = target more compressed).

Table 14 shows the smoother and lower-rank patterns of the EMA target encoder. The target encoder produces higher adjacent-timestep cosine similarity on all 10 datasets (mean 0.860 → 0.984) and lower effective rank on 9 of 10 datasets (mean 74.9 → 10.3, mean per-dataset compression ratio = 11.0×). The single exception in effective rank is SpokenArabicDigits, where the online encoder already exhibits a low rank (8), reflecting the intrinsic low-dimensionality of multivariate speech; the smoothing pattern still holds in this case (0.794 → 0.999). On the anomaly datasets (KPI and Yahoo), both encoders exhibited relatively low effective ranks compared to the forecasting datasets, consistent with the longer sequence length and per-timestep regularity of the anomaly signal; the directional gap (target < online) was preserved.

Geometrically, these properties align with task structure: classification benefits from a higher-rank representation that preserves fine-grained discriminative dimensions for sample-level decision boundaries, which the online encoder retains because it has not been smoothed by EMA averaging, whereas per-timestep regression (forecasting and anomaly detection) benefits from smooth trajectories along which nearby timesteps are tightly coupled and distance-based scoring functions (Ridge regression, k-NN, Mahalanobis) operate on a small number of dominant temporal modes.

To test whether this asymmetry is reducible to simple weight averaging within a single-encoder framework, CF-JEPA's EMA target was compared with a Stochastic Weight Averaging (SWA) baseline trained as a single encoder, measuring the same geometric properties (effective rank and adjacent-timestep cosine similarity) used in Table 14.

Across the same 10 datasets (Table 15), SWA produces representations virtually identical to the online encoder ($\Delta r_{\text{eff}} = +1.0$, $\Delta\overline{cos_t} = -0.002$), whereas the EMA target encoder shows dramatically different geometric properties ($\Delta r_{\text{eff}} = -64.6$, $\Delta\overline{cos_t} = +0.124$). This demonstrates that dual-encoder dynamics introduce structural smoothing and rank compression, beyond what simple weight averaging in a single-encoder framework can produce.

The key distinction lies in how the EMA target participates in the loss: whereas SWA averages weights post-hoc and is never observed by the optimizer, CF-JEPA's EMA target is the prediction target in the JEPA loss, so the online encoder is optimized to match representations produced by an EMA of its own past weights. This bootstrap loop induces self-consistency across the EMA window, suppressing directions of variation along which the online network has not yet converged and yielding slow-varying, lower-rank representations. SWA, lacking this loss-coupling, inherits the online encoder's full rank and roughness.

**Table 15.** Stochastic Weight Averaging (SWA) control for the dual-encoder mechanism.

| Encoder | Mean $r_{\mathrm{eff}}$ | Mean $\overline{cos}_t$ | Mean $\overline{cos}_{\mathrm{bin}}$ |
|---|---|---|---|
| Online encoder | 74.9 | 0.8599 | 0.9251 |
| SWA (single encoder) | 75.9 | 0.8579 | 0.9276 |
| EMA target encoder | 10.3 | 0.9837 | 0.9585 |

Table 15. Comparison of the online encoder, an SWA baseline (single encoder; weights averaged over the last 50% of training epochs, learning rate schedule unchanged), and the EMA target encoder. The same diagnostics, datasets, and seed as in Table 14 are used.

The two encoders are not redundant copies; they are complementary deployment modes jointly produced from a single training run. Online routing for classification, forecasting targets, and anomaly detection is a specific design contribution of the CF-JEPA downstream protocol. This is enabled by the dual-encoder dynamics of the JEPA framework. Whether other EMA-based dual-encoder methods (BYOL and SimSiam) allow similar asymmetric routing remains an open question. This is the strongest empirical evidence supporting the asymmetric utilization claim that distinguishes CF-JEPA from previous time-series JEPA variants.

## 5. Implications of the research

The contributions of CF-JEPA extend beyond a single methodological advance. The findings of this study have theoretical, methodological, and practical implications for time-series self-supervised learning, with relevance to the broader SSL community.

### *5.1 Theoretical implications*

A counterexample to the assumption that masking is the default for time-series SSL. This paper empirically demonstrates that masking is not a necessary component of time-series JEPA, and is structurally ill-suited to temporally continuous signals. The assumption that the success of BERT, MAE, or JEPA on images is automatically transfers to time series has gone largely unchallenged. Our results reinforce a more general principle: the geometric structure of the data (spatial redundancy in images vs. temporal continuity in time series) should govern the choice of self-supervised signal. This suggests that future SSL designs in other domains (such as graphs, multimodal data, and biosignals) should select learning signals tailored to the structural properties of their data.

The discovery and generality of dual-encoder asymmetry. We show that EMA-based dual-encoder training is not merely a stabilization device but also a mechanism for producing two representation modes at no extra cost. This raises a hypothesis applicable to other SSL frameworks employing EMA targets—BYOL, SimSiam, MoCo, I-JEPA, and V-JEPA—namely, that the online/target asymmetry may be a general property of EMA-based SSL rather than an artifact of time-series JEPA. Framing the EMA as a low-pass filter in parameter space provides a conceptual starting point for investigating this generalization in computer vision and NLP.

The task-encoder matching principle is based on effective rank and temporal smoothness. We determined that effective rank and adjacent timestep cosine similarity systematically matched the downstream tasks (Table 14, 15). This indicates that "good representations" are not monolithic; different geometric properties favor different task inductive biases. Future SSL evaluations may benefit from restructuring the alignment between task inductive bias and representation geometry, rather than relying on a single downstream metric.

### *5.2 Methodological implications*

Reconsidering predictor capacity and placement. The CF-JEPA ablation showed that a linear near-identity predictor matched or outperformed the MLP/Transformer predictors (Table 11), in contrast to the deep predictors used in I-JEPA and V-JEPA. This suggests that for time-series SSL, it is more effective to allocate capacity to the encoder rather than to the predictor, providing a design guideline that improves both efficiency and reproducibility.

Explicit recognition of backbone–objective compatibility. Table 12 shows that the TS2Vec dilated convolution encoder, designed for contrastive objectives, performs 6.3% worse under our forward-predictive objective, even when capacity is matched. This indicates that the encoder architecture and learning objective are not independent design choices but must be co-designed as a compatible pair. Future time-series SSL research should avoid adopting state-of-the-art encoders without considering whether their inductive biases align with the learning objective.

Need for unified evaluation protocols. By rerunning 11 baselines under identical epoch budgets, batch sizes, and multi-seed protocols, this study partially resolves the inconsistencies in cross-paper comparisons that affect time-series SSL. This demonstrates that the field must standardize from iteration-based to epoch-based training budgets, controlling for dataset size effects to enable fair comparisons.

### *5.3 Practical implications*

Efficient self-supervised learning for industrial and embedded time-series applications. In industrial settings, such as manufacturing, energy, finance, and healthcare, where time-series data are abundant but labels are scarce, CF-JEPA supports three core tasks—classification, forecasting, and anomaly detection—using a single pretrained model. Obtaining two encoders from one training run at no additional cost is particularly valuable under resource constraints.

Extensibility for edge and embedded environments. The depthwise-pointwise-factorized multi-scale DWConv encoder delivers equal or better performance than Transformer baselines with substantially fewer parameters, enabling real-time SSL inference on factory PLCs, IoT gateways, and wearable medical devices. Further compression via smaller representation dimensions and shallower architectures, combined with quantization and pruning, can enable deployment in even more constrained settings.

Integrated multitask monitoring system. Industrial time-series monitoring typically requires three concurrent tasks: classifying normal/abnormal values, predicting future values, and detecting anomalous time steps. CF-JEPA's dual-encoder structure supports all three modes from a single pretraining run, reducing operational costs for model management, deployment, and maintenance in production systems.

Applicability to artificial intelligence manufacturing and smart factory environments. This research applies time-series learning directly to manufacturing process data, including vibration, temperature, and current signals. The ability to perform defect classification, lifespan prediction, and early failure detection with a single label-free pretrained model provides foundational capabilities for digital data factory construction and autonomous factory operation.

## 6. Conclusion

CF-JEPA was introduced in this study as a self-supervised representation learning framework for time-series data, replacing the masking strategy of the JEPA family with multi-horizon forward prediction. By using random crops as views and predicting future representations in the forward temporal direction, CF-JEPA learns abstract representations in the embedding space without requiring data augmentation or contrastive pairs.

A key finding from analyzing the training dynamics of CF-JEPA is that the online and EMA target encoders, jointly produced from a single training run, develop distinct representational properties suited to different downstream tasks. This asymmetry is exploited by routing classification to the online encoder, and forecasting and anomaly detection to the EMA target encoder. Ablation analysis shows that this asymmetric utilization yields a 27% reduction in multivariate forecasting MSE compared to using the online encoder of the same trained model for forecasting, providing strong empirical evidence in support of the dual-encoder design at no additional training cost. Across 126 UCR and 26 UEA classification datasets, eight forecasting benchmarks, and KPI/Yahoo anomaly detection, among self-supervised time-series representation learning methods, CF-JEPA achieved the highest Avg Acc and Avg Rank on UCR and UEA classification, with the top three baselines within the Nemenyi critical difference, and remained competitive in forecasting and anomaly detection (top two Avg Rank on univariate forecasting and k-NN-scored anomaly detection, third on multivariate forecasting).

There are several avenues for future research. Pursuing more compact self-supervised representation learning with a smaller encoder size and lower representation dimensionality would extend CF-JEPA's applicability to resource-constrained settings, such as edge devices and embedded time-series systems. In addition, a deeper investigation into the mechanism by which the EMA target encoder produces representations beneficial for dense output tasks may broaden applicability to other self-supervised frameworks beyond time-series data.

## CRediT authorship contribution statement

**Jaehoon Lee**: Conceptualization, Methodology, Software, Validation, Formal analysis, Visualization, Writing – original draft. **Sunghyun Sim**: Conceptualization, Methodology, Supervision, Writing – review & editing, Project administration, Funding acquisition.

## Declaration of competing interest

The authors declare that they have no competing financial interests or personal relationships that may have influenced the work reported in this study.

**Data availability**

The following datasets used in this study are publicly available: UCR Time Series Classification Archive [31], UEA Multivariate Time Series Classification Archive [32], ETT benchmark [33], KPI anomaly detection dataset [34], and Yahoo Webscope S5 dataset [35]. The source code and pretrained models for CF-JEPA will be made publicly available at https://github.com/WDSLab/CF-JEPA.

**Acknowledgments**

This work was supported by the National Research Foundation of Korea (NRF) grant funded by the Korean government (MSIT) (No.RS-2023-00218913) and was supported by Basic Science Research Program through the National Research Foundation of Korea(NRF) funded by the Ministry of Education(No.RS-2025-25396743)